\documentclass[conference]{IEEEtran}
\IEEEoverridecommandlockouts
\usepackage{cite}
\usepackage{amsmath,amssymb,amsfonts}
\usepackage{booktabs}
\usepackage{graphicx}
\usepackage{textcomp}
\usepackage{xcolor}
\usepackage{url}
\def\BibTeX{{\rm B\kern-.05em{\sc i\kern-.025em b}\kern-.08em
    T\kern-.1667em\lower.7ex\hbox{E}\kern-.125emX}}
\usepackage{listings}
\usepackage{xcolor}

\usepackage{arydshln}
\definecolor{backcolour}{rgb}{0.95,0.95,0.92}
\lstdefinestyle{mystyle}{
    backgroundcolor=\color{backcolour},   
    basicstyle=\ttfamily\footnotesize,
    breaklines=true,
    numbers=left,
}
\begin{document}

\title{Cross-Attention Calibrated Deduplication for Retrieval-Augmented Generation System\\
}

\author{
  \IEEEauthorblockN{
    Phuong Le Huy\IEEEauthorrefmark{1}, 
    Nam H. Nguyen\IEEEauthorrefmark{1}, and 
    Quan V. Dang\IEEEauthorrefmark{1},\IEEEauthorrefmark{2}
  }
  \IEEEauthorblockA{
    \IEEEauthorrefmark{1}Full Stack Data Science\\
    \IEEEauthorrefmark{2}Department of Computer Science, University College London\\
    Emails: phuonglehuy172k@gmail.com, namnqs12@gmail.com,  dangvanquan.nd@gmail.com
  }
}

\maketitle

\begin{abstract}
Common chunking strategies in Retrieval-Augmented Generation (RAG) systems often create redundant chunks. These redundant chunks make the vector database bigger and slow down retrieval. A common fix is cosine-similarity thresholding. This method reduces each chunk to a single vector, then compares vectors using a similarity score. But a single vector can lose the fine-grained, token-level detail needed to tell a true duplicate apart from a chunk that just shares the same topic. We propose Cross-Attention Calibrated Deduplication (CACD). CACD checks each new chunk against an in-memory pool of chunks already kept, using a cross-encoder instead of a single pooled vector. This keeps token-level detail all the way to the final comparison. CACD combines three parts: the cross-encoder comparison itself, a New Information Score (NIS) that measures how much of a chunk is not explained by a candidate already kept, and a majority vote across several candidates rather than a single best match. NIS is calculated from the attention entropy of the cross-encoder. We tested CACD against five existing filtering methods, nine chunking strategies, and 18 configurations, all on the full SQuAD~1.1 validation set. In our experiments, CACD removes 9.75\% of chunks on average. This drop rate is close to other semantic-level methods, and much higher than exact-match filters, which barely remove anything. In these experiments, CACD also processes each configuration in 51.0 seconds on average, about 27\% faster than the strongest baseline, NERExact (69.6s), and about 7$\times$ faster than cosine-similarity filtering (356.7s). These results come from a single dataset, so we present them as an early comparison, not a general claim. Code for the baseline evaluation and for CACD is available at \url{https://github.com/lehuyphuong/rag_bench} and \url{https://github.com/lehuyphuong/cacd_dedup}.
\end{abstract}


\section{Introduction}

Retrieval-Augmented Generation (RAG) is a way to make a language model answer questions using information it was not trained on~\cite{lewis2020retrieval}. Instead of relying only on what the model memorized during training, a RAG system keeps an external store of documents, retrieves the passages most relevant to a question, and gives those passages to the model as context before it generates an answer. This reduces hallucination and lets the system stay up to date without retraining the model itself.

\begin{figure}[!t]
\centering
\includegraphics[width=\columnwidth]{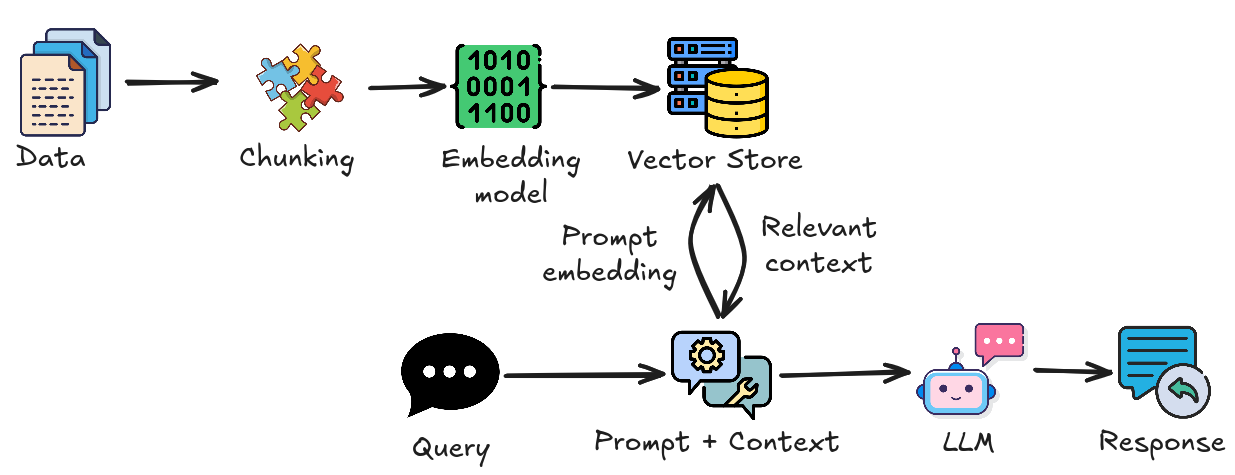}
\caption{A typical RAG pipeline. Documents are chunked, embedded, and stored in a vector store; at query time, the query is embedded, relevant chunks are retrieved, and the LLM generates a response from the combined prompt and context.}
\label{fig:RAG_pipeline}
\end{figure}

A RAG system is built from two connected flows. As shown in Figure 1, the first flow prepares the document collection for retrieval: raw data is split into smaller pieces, called chunks, by a chunking step, and each chunk is turned into a vector representation by an embedding model; these vectors are stored in a vector store. The second flow handles each incoming question: the query is embedded and sent to the vector store, which returns the most relevant chunks as context; this context is combined with the original query into a prompt, and a large language model (LLM) reads the prompt and produces the final response\cite{designveloper2025rag}.

As new documents keep being added to the vector store over time, chunking strategies often produce chunks that overlap or repeat the same information, sometimes from the same document and sometimes across different documents \cite{lee2022deduplicating} \cite{shah2023r2d2}. This redundancy grows the vector store without adding new information, slows down retrieval, and can even hurt answer quality if the retrieved context is full of repeated content instead of diverse, relevant information. Deciding whether a new chunk is a genuine duplicate of something already in the vector store, without discarding a chunk that only shares a topic, is the central problem this paper addresses.

The main contributions of this paper are:
\begin{itemize}
  \item Cross-Attention Calibrated Deduplication (CACD), a filtering method with three parts working together: a cross-encoder that compares each new chunk against the full, persistently growing index instead of pooled similarity vectors; a New Information Score (NIS), derived from the entropy of the cross-encoder's attention matrix, that scores how much of a chunk is not explained by a given candidate; and a majority vote across several retrieved candidates, so that one misleading nearest neighbor cannot flip the outcome on its own.
  \item An evaluation of five existing filtering methods that had not previously been benchmarked against one another, alongside CACD, across nine chunking strategies and eighteen configurations on the SQuAD 1.1 validation set.
\end{itemize}

The rest of this paper is organized as follows. Section II reviews related work on chunking strategies and existing filtering methods. Section III describes CACD in detail: the retrieval, scoring, and decision stages, the guards that protect against known failure cases, and the decision thresholds used. Section IV presents the experimental setup and results, comparing CACD against the five baseline methods and breaking results down by chunking strategy. Section V concludes and discusses current limitations.

\section{Related Work}

\subsection{Chunking Strategies}\label{sec:related-work-chunking}
\begin{figure}[!t]
\centering
\includegraphics[width=\columnwidth]{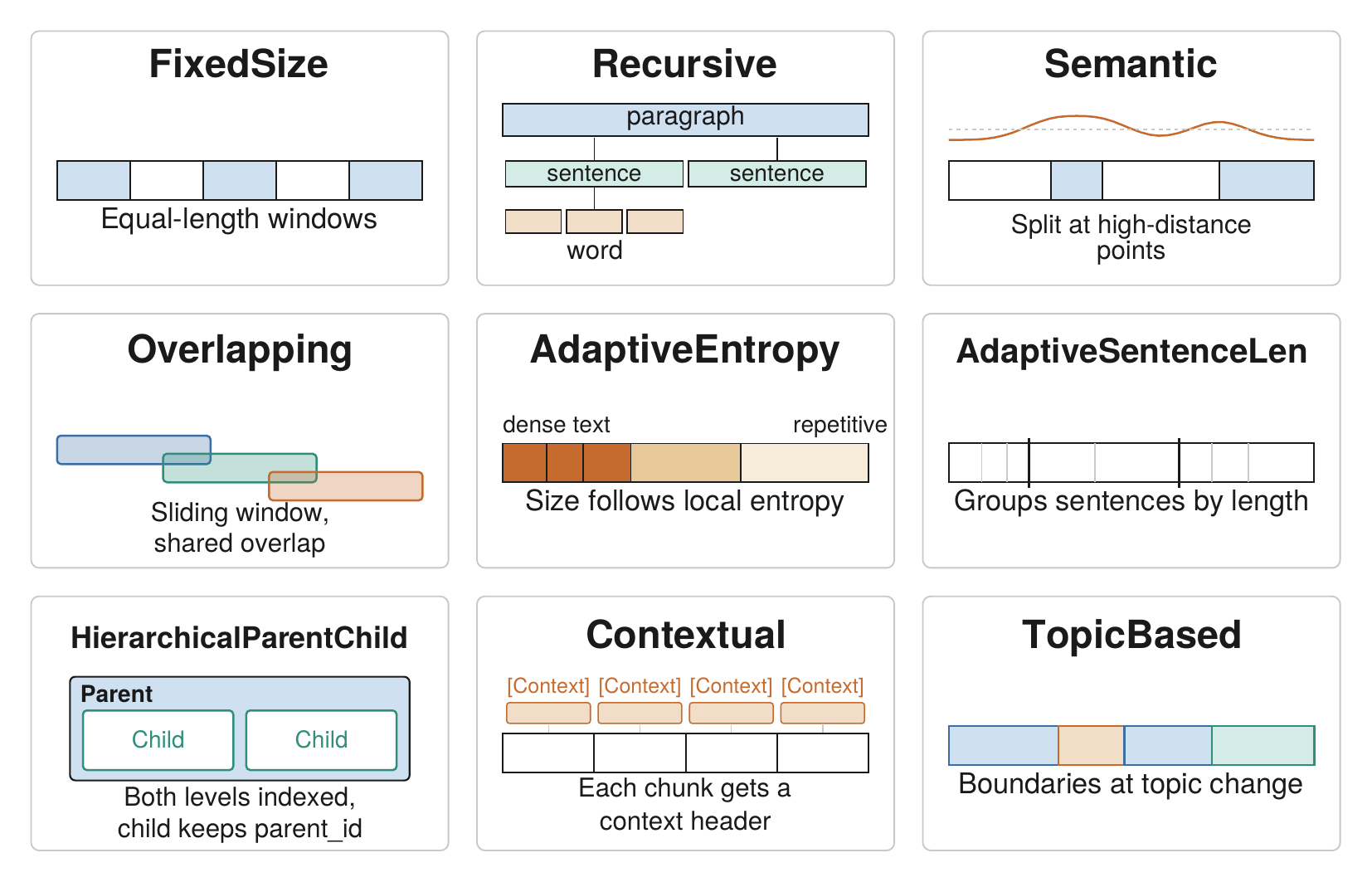}
\caption{Schematic overview of the nine chunking strategies compared in this work. Each panel shows how one document (horizontal bar) is split into chunks under that strategy's mechanism.}
\label{fig:chunking-strategies}
\end{figure}

RAG systems can split documents in many ways before indexing. Figure~\ref{fig:chunking-strategies} shows the nine strategies compared here.

\begin{itemize}
  \item \textbf{FixedSize} splits text into equal-length windows, ignoring sentence or paragraph breaks~\cite{bhat2025rethinking}.
  \item \textbf{Recursive} splits at paragraph breaks first, then sentence breaks, then word breaks, using a smaller split only when a larger one does not fit~\cite{kreileder2026evaluating}.
  \item \textbf{Semantic} places a boundary between two nearby sentences when the distance between their embeddings is high, keeping related sentences together~\cite{qu2025semantic}.
  \item \textbf{Overlapping} uses a sliding window with overlap between chunks, so text near an edge can appear in more than one chunk~\cite{smigielski2026chunking}.
  \item \textbf{AdaptiveEntropy} uses smaller chunks where local text entropy is high (information-dense) and larger chunks where it is low (repetitive)~\cite{kondapalli2026queryaware}.
  \item \textbf{AdaptiveSentenceLen} groups more short sentences or fewer long ones per chunk, based on the average sentence length nearby~\cite{kondapalli2026queryaware}.
  \item \textbf{HierarchicalParentChild} indexes a larger parent chunk together with its smaller child chunks, each child linked back to its parent~\cite{elchafei2026hrag}.
  \item \textbf{Contextual} adds a short header, such as the document title, to each chunk before embedding, so the embedding also carries some context from the document~\cite{singh2025reconstructing}.
  \item \textbf{TopicBased} groups sentence embeddings with k-means and joins nearby sentences from the same group into one chunk, placing a boundary where the topic changes~\cite{hadawle2025chunking}.
\end{itemize}

\subsection{Chunk Deduplication Techniques}\label{sec:related-work-filtering}
\begin{figure}[!t]
\centering
\includegraphics[width=\columnwidth]{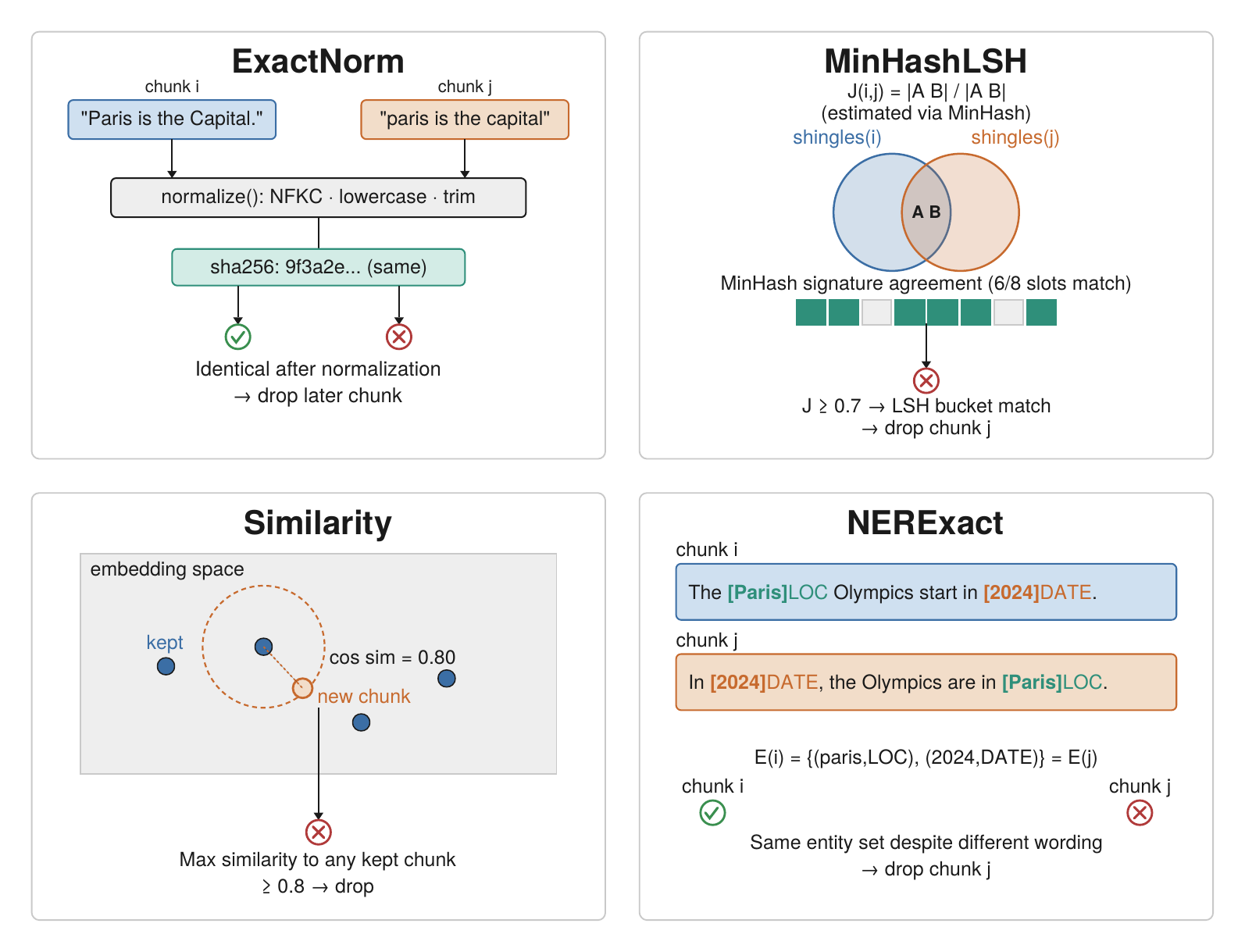}
\caption{Schematic overview of the four baseline deduplication filters compared against CACD. ExactNorm and MinHashLSH compare lexical or character-level similarity; Similarity compares dense embedding cosine distance; NERExact compares named-entity set equality.}
\label{fig:dedup-filters}
\end{figure}

Alongside chunking strategies, several filtering methods have been used to remove redundant chunks before indexing~\cite{berdyugina2026reducing}. Figure~\ref{fig:dedup-filters} summarizes the four baseline methods compared against CACD in this work; each appears to have a different blind spot.

\begin{itemize}
  \item \textbf{ExactNorm} removes a chunk if its text, after normalization (case-folding and whitespace collapsing), exactly matches an already-kept chunk~\cite{berdyugina2026reducing}. Because it requires an exact match, it may not catch chunks that convey the same information with even minor wording differences.
  \item \textbf{MinHashLSH} estimates Jaccard similarity between chunks from character-level shingle sets and removes a chunk once its estimated similarity to an already-kept chunk crosses a threshold~\cite{berdyugina2026reducing}. Since it operates on lexical overlap, it may miss paraphrased duplicates and, depending on the shingle size and threshold chosen, may also flag lexically similar but topically distinct chunks.
  \item \textbf{Similarity} embeds each chunk into a single pooled vector and removes it if its cosine similarity to any already-kept chunk exceeds a threshold~\cite{berdyugina2026reducing}. Collapsing a chunk to one vector can make it difficult to tell a genuine duplicate apart from a chunk that only shares a topic or vocabulary, which may lead to removing chunks that are not actually redundant.
  \item \textbf{NERExact} extracts the named entities in each chunk and removes a later chunk if its entity set exactly matches an already-kept chunk's~\cite{berdyugina2026reducing}. This signal depends entirely on named entities, so it may leave redundancy in entity-sparse text unaddressed, and two chunks that mention the same entities can still be treated as duplicates even if they describe different information about those entities.
\end{itemize}

These four methods and CACD are compared under the same experimental setup in Section~\ref{sec:setup}.

\section{Methods}

\subsection{Overview}\label{sec:overview}
Cross-Attention Calibrated Deduplication (CACD) processes chunks one at a time as a document collection is ingested. For each new chunk $A$, CACD decides whether to insert it into a persistent index $I$ or discard it as redundant, checking it against every chunk kept so far in the same ingestion run, not only the chunks in the current batch.

\begin{figure}[!t]
\centering
\includegraphics[width=\columnwidth]{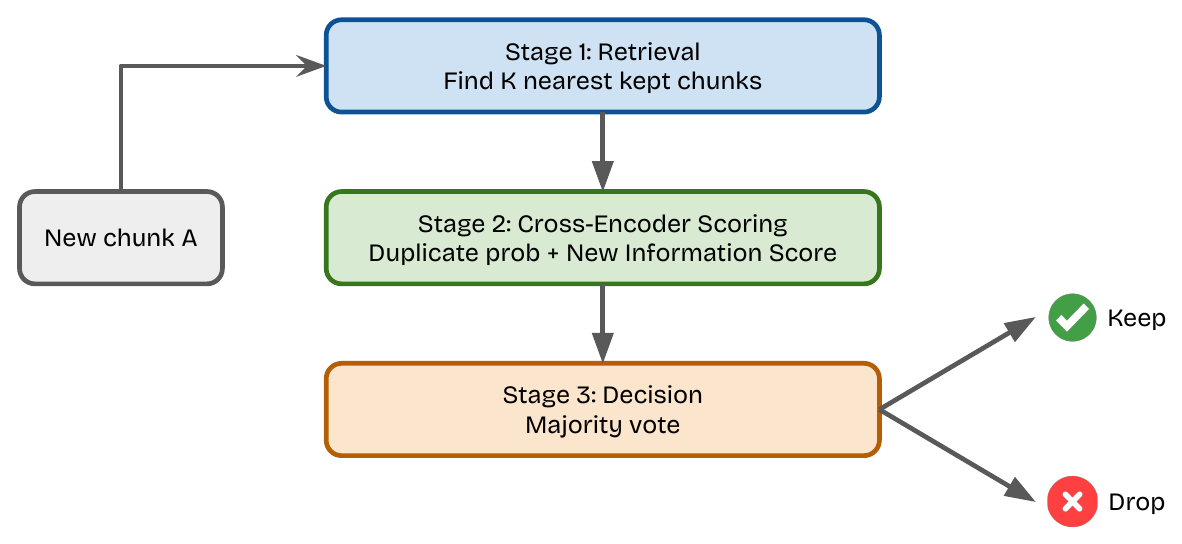}
\caption{CACD's three-stage pipeline. A new chunk is scored against the K nearest chunks already kept, and a majority vote across those K candidates decides whether to keep or drop it.}
\label{fig:cacd-overview}
\end{figure}

The pipeline has three stages, shown in Figure~\ref{fig:cacd-overview}. Stage 1 retrieves the $K$ chunks in $I$ closest to $A$. Stage 2 scores $A$ against each of these $K$ candidates with a cross-encoder, producing a duplicate probability and a New Information Score (NIS). Stage 3 turns these per-candidate scores into one \textsc{Keep}/\textsc{Drop} decision for $A$ through a majority vote, subject to the guards in Section~\ref{sec:guards}. A chunk voted \textsc{Drop} is discarded.

\subsection{Stage 1: Retrieval}\label{sec:stage1}
\begin{figure}[!t]
\centering
\includegraphics[width=\columnwidth]{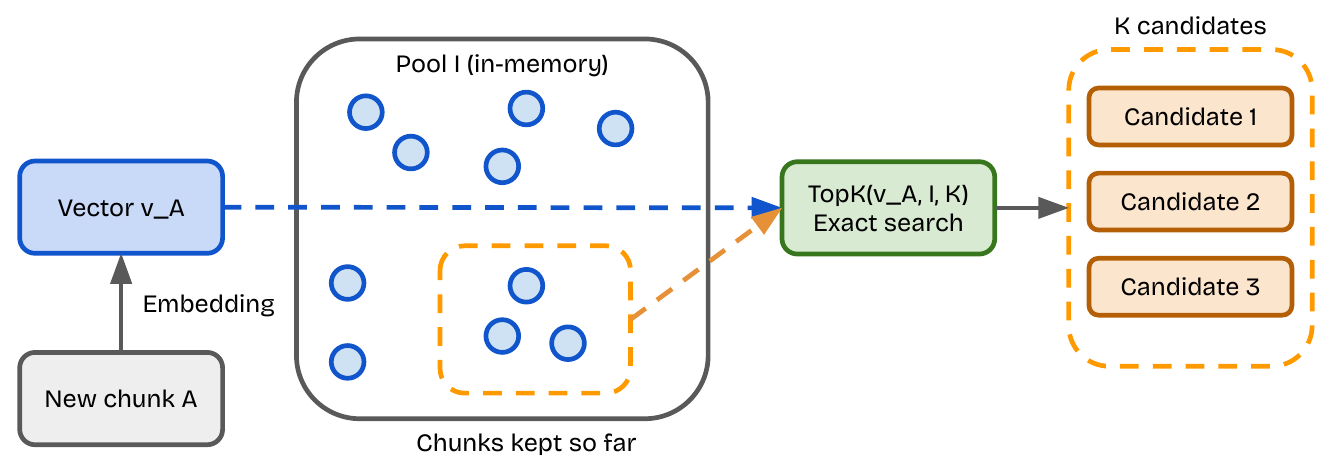}
\caption{Stage 1: the new chunk is embedded, then compared against every chunk already kept in the in-memory pool to retrieve the K nearest candidates.}
\label{fig:stage1-detail}
\end{figure}

$A$ is embedded into a vector $v_A$, and the $K$ nearest chunks ~\cite{cunningham2020knn} ~\cite{mahon2025kstarmeans} already in $I$ are retrieved by an exact search over $I$ held in memory (Figure~\ref{fig:stage1-detail}): $I$ is a single growing matrix of embeddings, and retrieval is one matrix--vector product followed by picking the $K$ largest scores.

This search is exact, not approximate, so it always finds the true $K$ nearest chunks under cosine similarity, unlike an approximate index such as HNSW \cite{malkov2020efficient}, at the cost of scanning the whole pool per query rather than a faster approximate lookup. At the data sizes used in this paper (a few thousand to roughly ten thousand chunks per configuration), this was still fast in wall-clock terms in our experiments: an earlier version of Stage~1 built on an external vector store was in practice slowed down mostly by per-query storage overhead, which the in-memory version removes. If $I$ is still empty, $A$ is kept automatically, since there is nothing yet to compare it against.

\subsection{Stage 2: Cross-Encoder Scoring}\label{sec:scoring}
\begin{figure}[!t]
\centering
\includegraphics[width=\columnwidth]{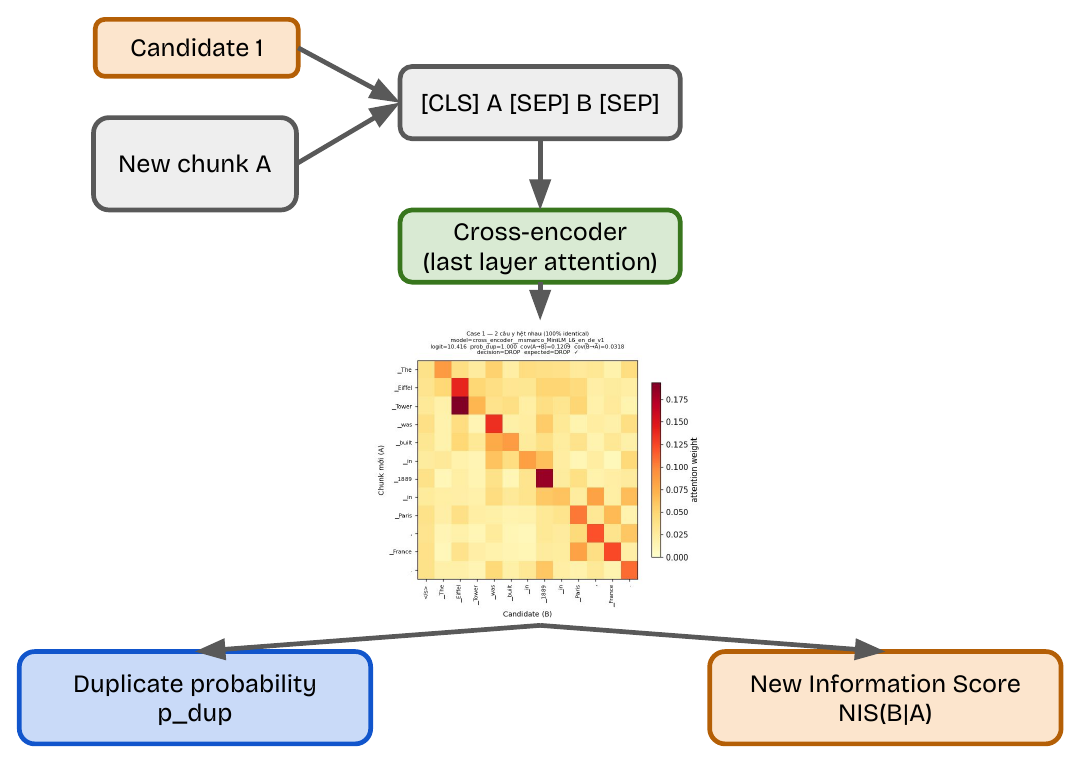}
\caption{Stage 2: chunk A and a candidate are encoded together; an example attention matrix is shown, from which both a duplicate probability and the New Information Score are computed.}
\label{fig:stage2-detail}
\end{figure}

Each candidate pair $(A, B)$ is jointly encoded by a pretrained cross-encoder \cite{vast2025understanding}\cite{ananthakrishnan2025crossencoders} as one sequence, $[\text{CLS}]\ A\ [\text{SEP}]\ B\ [\text{SEP}]$, in a single forward pass (Figure~\ref{fig:stage2-detail}). This produces a duplicate probability $p_{\text{dup}} \in [0,1]$ and the model's final-layer attention matrix, averaged across attention heads.

The New Information Score (NIS) uses the part of that attention matrix showing how tokens of $B$ attend back to tokens of $A$. For each token $j$ of $B$, its attention over $A$ is turned into a probability distribution $p(\cdot \mid j)$ over the tokens of $A$, and its Shannon entropy \cite{vajapeyam2014entropy} is computed:
\begin{equation}
H(j) = -\sum_{i=1}^{|A|} p(i \mid j) \log p(i \mid j),
\label{eq:entropy}
\end{equation}
where $|A|$ is the number of tokens in $A$. A low $H(j)$ means token $j$'s attention concentrates on a few tokens of $A$, so $A$ explains $j$; a high $H(j)$ means attention spreads evenly across $A$, so $j$ carries information $A$ does not have. $H(j)$ is normalized by $\log|A|$, the value it takes when attention is exactly uniform over $A$, and averaged over all $|B|$ tokens $j$ of $B$ to give
\begin{equation}
\mathrm{NIS}(B \mid A) = \frac{1}{|B|}\sum_{j=1}^{|B|} \frac{H(j)}{\log |A|} \in [0,1].
\label{eq:nis}
\end{equation}
$\mathrm{NIS}(B \mid A) \to 0$ means $B$ is fully explained by $A$ (redundant); $\mathrm{NIS}(B \mid A) \to 1$ means $B$ is largely unaccounted for by $A$.

\begin{figure}[!t]
\centering
\includegraphics[width=\columnwidth]{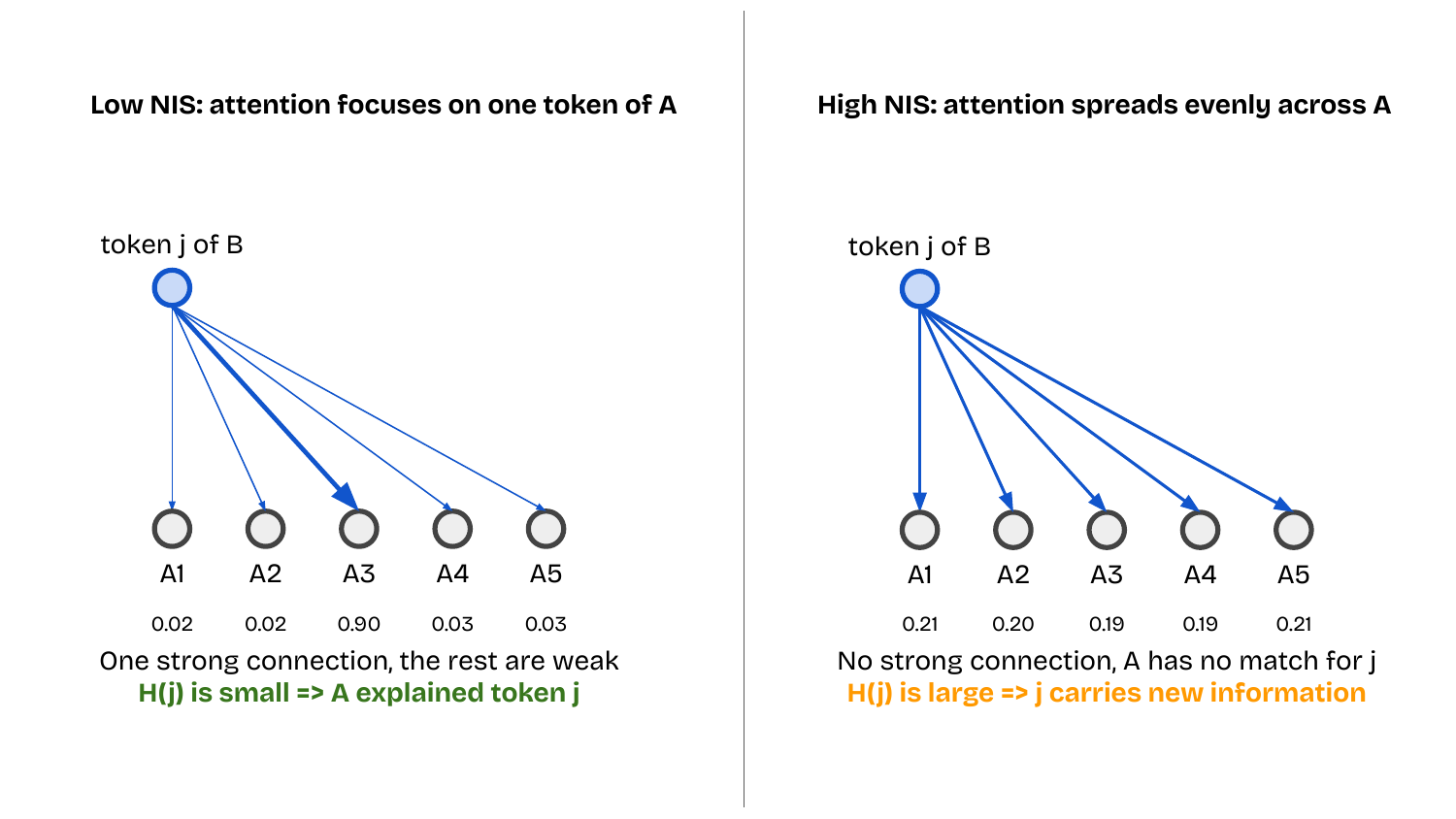}
\caption{How a single token $j$ of $B$ attends to the tokens of $A$ in the two extreme cases behind NIS. When attention concentrates on one token of $A$ (left), $H(j)$ is small and $A$ is treated as explaining $j$. When attention spreads evenly across $A$ (right), $H(j)$ is large and $j$ is treated as carrying information $A$ does not have.}
\label{fig:nis-visualized}
\end{figure}

Figure~\ref{fig:nis-visualized} illustrates both cases for a single token $j$ of $B$. When one token of $A$ receives most of the attention, $H(j)$ stays small, so $j$ is treated as explained by $A$. When attention spreads evenly across all tokens of $A$ instead, $H(j)$ grows toward its maximum, so $j$ is treated as carrying information $A$ does not have.

To check whether $p_{\text{dup}}$, the cross-encoder's own duplicate probability, gives a reasonable signal on its own, chunk pairs were built with a known, exact overlap level, from 100\% down to 0\% in steps of 10\%. One chunk stayed fixed. The other was built from paraphrases (not copies) of a set fraction of the first chunk's sentences, with the rest replaced by unrelated sentences. This was done for two different topic pairs, and the results were averaged.

Table~\ref{tab:nis-vs-cosine} shows cosine similarity and $p_{\text{dup}}$ at each overlap level. $p_{\text{dup}}$ follows roughly the same downward trend as cosine similarity as overlap decreases, though the pattern is not fully clean: it flattens at a couple of points (40\% and 30\% overlap give nearly the same value), and at 10\% overlap it sits slightly above its value at 20\%. Overall, $p_{\text{dup}}$ lands close to cosine similarity rather than clearly ahead of or behind it, which is a reasonable result for a signal used entirely on its own, without the guards or the majority vote described in Sections~\ref{sec:guards} and~\ref{sec:decision}.

To do better than $p_{\text{dup}}$ alone, CACD does not stop here: it combines $p_{\text{dup}}$ with NIS and a majority vote across several candidates, as described in Section~\ref{sec:decision}. Table~\ref{tab:main-results} and Table~\ref{tab:per-strategy} report how the full CACD pipeline performs against the baseline filters on the SQuAD 1.1 validation set.

\begin{table}[!t]
\centering
\caption{Cosine similarity vs. $p_{\text{dup}}$ across a semantic-overlap range, averaged over two topic pairs.}
\label{tab:nis-vs-cosine}
\scriptsize
\begin{tabular}{lcc}
\toprule
\textbf{Overlap \%} & \textbf{Cosine} & \textbf{$p_{\text{dup}}$} \\
\midrule
100 & 0.888 & 0.817 \\
90  & 0.864 & 0.693 \\
80  & 0.796 & 0.673 \\
70  & 0.780 & 0.631 \\
60  & 0.738 & 0.581 \\
50  & 0.719 & 0.519 \\
40  & 0.676 & 0.478 \\
30  & 0.601 & 0.478 \\
20  & 0.553 & 0.421 \\
10  & 0.193 & 0.293 \\
0   & 0.036 & 0.133 \\
\bottomrule
\end{tabular}
\end{table}

\subsection{Guards}\label{sec:guards}
Three conditions adjust the decision around scoring.

The parent-child guard excludes a candidate $B$ from scoring against $A$ when hierarchical chunking links them as parent and child, since their overlap is intentional, not duplication; sibling children of the same parent are still scored normally, since they may genuinely duplicate each other.

The header guard strips any prepended \texttt{[Context: \ldots]} header from both $A$ and $B$ before scoring only, so a header shared by every chunk of a document does not dominate the comparison; the stored text keeps the header.

The length-aware guard protects a chunk longer than 300 characters from being dropped by a high $p_{\text{dup}}$ or a low NIS alone, on the assumption that a longer chunk is more likely to carry unique information, unless its NIS against the deciding candidate falls below a floor of 0.3, indicating the candidate explains it too thoroughly for length alone to justify keeping it.

\subsection{Stage 3: Decision}\label{sec:decision}
The two probability thresholds used below, $\tau_{\text{high}}$ and $\tau_{\text{low}}$, and the NIS threshold $\tau_{\text{NIS}}$, are not hand-picked: $\tau_{\text{high}}$ and $\tau_{\text{low}}$ come from a cost-sensitive cutoff \cite{elkan2001foster} that balances the cost of wrongly dropping a non-duplicate chunk against the cost of wrongly keeping a genuine duplicate, which under the symmetric setting used in this paper's experiments gives $\tau_{\text{high}} = 0.8$ and $\tau_{\text{low}} = 0.2$; $\tau_{\text{NIS}} = 0.8$ is the midpoint of the normalized entropy scale from Eq.~\ref{eq:nis}, following the retention target used in SemDeDup \cite{abbas2023semdedup}.

\begin{figure}[!t]
\centering
\includegraphics[width=\columnwidth]{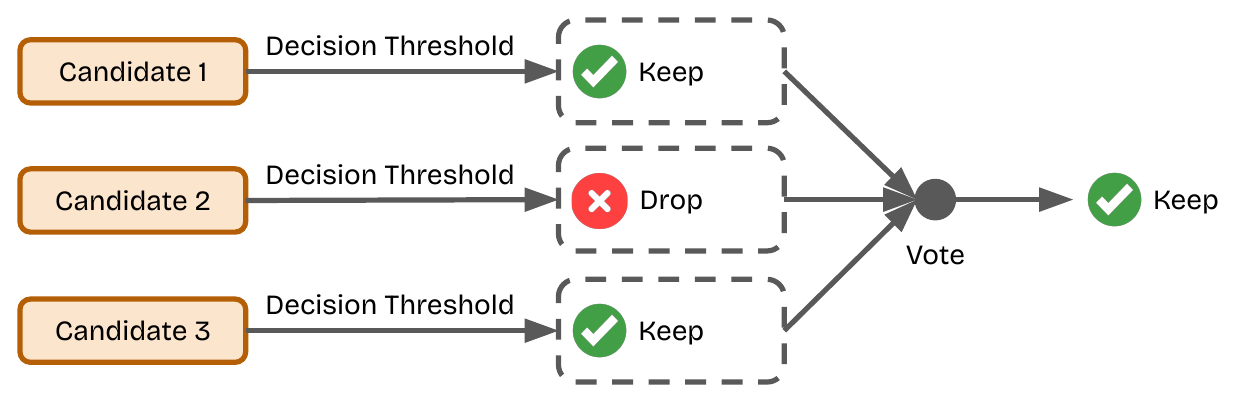}
\caption{Stage 3: each candidate is compared against the decision thresholds and casts a Keep or Drop vote; the majority across all votes decides the outcome for chunk A.}
\label{fig:stage3-detail}
\end{figure}

An earlier version of CACD based its decision on a single candidate, the one with the highest $p_{\text{dup}}$ among the $K$ retrieved. This made the outcome sensitive to a single retrieval error: if Stage 1 returned one misleadingly high-scoring but unrepresentative neighbor, $A$ could be dropped even though every other candidate indicated it was not a duplicate. CACD instead treats every valid candidate (one that passes the guards) as an independent vote ~\cite{louppe2014randomforests} ~\cite{abdoli2021bsac}, and requires a majority before committing to a decision ~\cite{yang2023contextualretrieval}, as shown in Figure~\ref{fig:stage3-detail}.

For each valid candidate $B$, a high $p_{\text{dup}}$ ($\geq \tau_{\text{high}}$) votes Drop, unless the length-aware guard protects $A$; a low $p_{\text{dup}}$ ($\leq \tau_{\text{low}}$) votes Keep; and in between, NIS decides, voting Keep when $B$ leaves enough of $A$ unexplained ($\geq \tau_{\text{NIS}}$) and Drop otherwise. Votes are tallied with early exit, stopping as soon as either side reaches a majority. If no majority is reached before every candidate has voted, CACD defaults to Keep, so an inconclusive vote never silently discards content.

The timing results in Section~\ref{sec:main-results} use a batched implementation that scores all $K$ candidates for a whole micro-batch of chunks in one cross-encoder pass before voting on any of them, which is faster on a GPU than voting candidate-by-candidate as described above. Both versions always reach the same decision.

\section{Results}\label{sec:setup}

\textbf{Setup.} Experiments were conducted on the full SQuAD 1.1 validation set \cite{rajpurkar2016squad} (2,067 passages, 10,570 question-answer pairs), across 18 chunking configurations built from nine chunking strategies at two size settings each. CACD was compared against five baselines (\textsc{NoFilter}, \textsc{ExactNorm}, \textsc{MinHashLSH}, \textsc{Similarity}, \textsc{NERExact}), all run on the same chunking outputs, embedding model, and evaluation steps, so differences in the results come from the filtering method itself.

\textbf{Configuration and metrics.} CACD used $\tau_{\text{NIS}}=0.8$, $K=5$, and symmetric costs ($c_{\text{FP}}=c_{\text{FN}}=1$), giving $\tau_{\text{high}}=0.8$ and $\tau_{\text{low}}=0.2$, the same values described in Section~\ref{sec:decision}. Precision, Recall, and IoU measure retrieval quality after filtering ~\cite{berdyugina2026reducing}; storage and ingestion time measure cost. GPU-accelerated embedding and FP16 cross-encoder inference were used, with Stage~1 retrieval run as an exact in-memory CPU search; these choices affect speed only, not which chunks any method keeps or drops.

\subsection{Main Results}\label{sec:main-results}

Table~\ref{tab:main-results} reports results averaged per chunking configuration across all 18 configurations. In this evaluation, CACD reaches the highest drop rate of any method (9.75\%, next closest is \textsc{Similarity} at 8.40\%) and the smallest index (27.13~MB, versus 27.38--30.64~MB for the rest). Ingestion time (51.01s) is close to \textsc{MinHashLSH} (50.74s) and far faster than the other semantic-level baselines, \textsc{Similarity} (356.70s) and \textsc{NERExact} (69.59s); only the two baselines that do little comparison work, \textsc{NoFilter} and \textsc{ExactNorm}, are faster.

Precision (0.3818) and IoU (0.3263) sit above \textsc{Similarity} and \textsc{NERExact} but below \textsc{NoFilter}, \textsc{ExactNorm}, and \textsc{MinHashLSH}, which is expected since those three remove very little content and so stay close to the \textsc{NoFilter} upper bound. Recall is the one area where CACD trails every baseline in this run (0.7001, at most 0.015 below \textsc{NERExact}), read as the cost of removing more content than any other method tested rather than a sign of weaker retrieval.

\begin{table}[!t]
\centering
\caption{Results averaged per chunking configuration (18 configurations), full SQuAD 1.1 validation set (2{,}067 documents, 10{,}570 questions). Storage and Time are per-configuration means, not totals. Best value per column is in \textbf{bold}, independent of method.}
\label{tab:main-results}
\scriptsize
\setlength{\tabcolsep}{3pt}
\begin{tabular}{lcccccc}
\toprule
\textbf{Method} & \textbf{Prec.} & \textbf{Rec.} & \textbf{IoU} & \textbf{MB} & \textbf{Time (s)} & \textbf{Drop \%} \\
\midrule
NoFilter            & \textbf{0.3924} & 0.7143 & \textbf{0.3364} & 30.64 & 32.16  & 0.00  \\
ExactNorm           & 0.3908 & 0.7152 & 0.3355 & 30.36 & \textbf{31.74}  & 0.63  \\
MinHashLSH (0.8)    & 0.3903 & \textbf{0.7153} & 0.3351 & 30.27 & 50.74  & 0.86  \\
Similarity (0.8)    & 0.3745 & 0.7122 & 0.3216 & 27.38 & 356.70 & 8.40  \\
NERExact            & 0.3798 & 0.7057 & 0.3256 & 28.29 & 69.59  & 6.37  \\
\hdashline
CACD                & 0.3818 & 0.7001 & 0.3263 & \textbf{27.13} & 51.01  & \textbf{9.75} \\
\bottomrule
\end{tabular}
\end{table}

Table~\ref{tab:per-strategy} breaks these results down by chunking strategy. \textsc{HierarchicalParentChild} has the highest Precision (0.5041) and IoU (0.4304), along with the highest drop rate (14.66\%) and the largest, slowest index, since its parent and child spans overlap on purpose, exactly the kind of overlap CACD is designed to catch. \textsc{AdaptiveSentenceLen} shows close to the opposite pattern: the highest Recall (0.8990), the smallest index, and the lowest drop rate (1.90\%), since its chunks are already short and largely non-redundant. \textsc{TopicBased} has the highest drop rate overall (15.55\%), suggesting its topic-grouped chunks carry more repeated content than simpler strategies.

Overall, in these experiments CACD's effect depends on how much real overlap a chunking strategy introduces: strategies that create structured or repeated content see the largest benefit, while strategies that already produce short, separate chunks see very little change.

\begin{table}[!t]
\centering
\caption{CACD results by chunking strategy, averaged over each strategy's two size configurations. Best value per column is in \textbf{bold}.}
\label{tab:per-strategy}
\scriptsize
\setlength{\tabcolsep}{3pt}
\begin{tabular}{lcccccc}
\toprule
\textbf{Strategy} & \textbf{Prec.} & \textbf{Rec.} & \textbf{IoU} & \textbf{MB} & \textbf{Time (s)} & \textbf{Drop \%} \\
\midrule
FixedSize                & 0.3841 & 0.6112 & 0.3086 & 27.09 & 51.50 & 8.38 \\
Recursive                & 0.4185 & 0.5684 & 0.3141 & 30.17 & 57.95 & 12.96 \\
Semantic                 & 0.3599 & 0.6783 & 0.3098 & 22.70 & 48.45 & 11.68 \\
Overlapping              & 0.4153 & 0.7850 & 0.3793 & 31.91 & 53.58 & 4.82 \\
AdaptiveEntropy          & 0.3394 & 0.7614 & 0.3110 & 18.87 & 31.70 & 6.89 \\
AdaptiveSentenceLen      & 0.2586 & \textbf{0.8990} & 0.2545 & \textbf{10.28} & \textbf{18.25} & 1.90 \\
HierarchicalParentChild  & \textbf{0.5041} & 0.7204 & \textbf{0.4304} & 49.94 & 91.86 & 14.66 \\
Contextual               & 0.3687 & 0.6622 & 0.3117 & 26.37 & 45.50 & 10.88 \\
TopicBased               & 0.3880 & 0.6147 & 0.3169 & 26.84 & 60.30 & \textbf{15.55} \\
\bottomrule
\end{tabular}
\end{table}

\section{Conclusion}\label{sec:conclusion}

This paper presented CACD, a chunk deduplication method that scores each new chunk against a persistently growing index using a cross-encoder, combining a calibrated duplicate probability with an attention-derived New Information Score (NIS) and a majority vote across several retrieved candidates rather than a single best match.

In this evaluation, across five filtering baselines, nine chunking strategies, and eighteen configurations on the full SQuAD 1.1 validation set, CACD reaches the highest drop rate of any method tested (9.75\%), while using less index storage than every baseline except Similarity and ingesting faster than both other semantic-level baselines (Similarity, NERExact). Precision and IoU land above Similarity and NERExact in this run, though below the three baselines that filter close to nothing and stay near the NoFilter upper bound by construction; Recall trails every other method by at most 0.0056 relative to the closest baseline (NERExact).

By chunking strategy, the benefit in these experiments is largest for HierarchicalParentChild, where deliberate parent/child overlap is exactly the case a pooled-vector score is most likely to mistake for redundancy, termed here false-redundancy collapse, and smallest for strategies like AdaptiveSentenceLen that already produce short, largely disjoint chunks.

However, three limitations are worth considering:
\begin{itemize}
  \item The cost ratio and NIS thresholds were chosen by comparing a handful of settings on this one dataset rather than through a principled calibration procedure, and whether a less aggressive threshold could narrow the Recall gap while keeping most of the storage and Precision/IoU benefit remains untested.
  \item A chunk voted DROP is discarded outright, including any small amount of content it might carry that is not explained by other kept chunks; this loss of partial information may help explain why CACD's Precision, Recall, and IoU trail NoFilter's upper bound in this evaluation.
  \item The probability of duplication depends on the quality of the cross-encoder model. By observing several experiments, we conclude that many cross-encoder models such as \textit{nli-deberta-v3-small, stsb-roberta-large, msmarco-MiniLM-L6-en-de-v1, qnli-distilroberta-base and qnli-electra-base} ~\cite{huggingface2026crossencoder} provide better outcomes. And, in order to increase recalls, precision and IoU metrics more, a fine-tune processing is highly recommended.
\end{itemize}

\end{document}